\documentclass[runningheads]{llncs}

\usepackage[T1]{fontenc}
\usepackage[square, comma, numbers, sort&compress, sectionbib]{natbib}
\renewcommand{\refname}{References}

\usepackage{amsmath, amssymb, bm}
\usepackage{graphicx,verbatim}
\usepackage[utf8]{inputenc} 
\usepackage{booktabs, multirow, makecell}
\usepackage{colortbl,xcolor}

\newcommand{\modelname}{CHASE}

\begin{document}
\title{Clinically-Grounded Hierarchical Classification for Consistent Chest X-ray Interpretation}
\titlerunning{CHASE}
\begin{comment}  %% Removed for anonymized MICCAI submission

\author{First Author\inst{1}\orcidID{0000-1111-2222-3333} \and
Second Author\inst{2,3}\orcidID{1111-2222-3333-4444} \and
Third Author\inst{3}\orcidID{2222--3333-4444-5555}}
%
\authorrunning{F. Author et al.}
\institute{Princeton University, Princeton NJ 08544, USA \and
Springer Heidelberg, Tiergartenstr. 17, 69121 Heidelberg, Germany
\email{lncs@springer.com}\\
\url{http://www.springer.com/gp/computer-science/lncs} \and
ABC Institute, Rupert-Karls-University Heidelberg, Heidelberg, Germany\\
\email{\{abc,lncs\}@uni-heidelberg.de}}

\end{comment}

% \author{Anonymized Authors}  %% Added for anonymized MICCAI submission
% \authorrunning{Anonymized Author et al.}
% \institute{Anonymized Affiliations \\
%     \email{email@anonymized.com}}

\author{Jong Hak Moon\thanks{Corresponding author}\orcidID{0000-0002-6708-3918} \and Minjun Kim\orcidID{0009-0000-3615-1220} \and Minjun Kim\orcidID{0009-0009-8429-5240}}
\index{Moon, Jong Hak}
\index{Kim, Minjun}
\index{Kim, Minjun}
\authorrunning{Moon et al.}
\institute{Yeji X, Seoul, South Korea\\
\email{\{jh.moon, jayden.kim, mj.kim\}@yejix.ai}}
  
\maketitle

\begin{abstract}

Accurate chest X-ray interpretation is inherently hierarchical.
Clinical decisions depend not only on what abnormality is present
but where it is situated, requiring reasoning from broad
anatomical systems down to specific pathological findings.
Yet existing automated systems largely treat this as a flat
classification problem, failing to capture inter-level
dependencies or enforce coherence between coarse and fine
predictions.
We propose \modelname{} (\textbf{C}lassification with
\textbf{H}ierarchical \textbf{A}nalysis and \textbf{S}tructured \textbf{E}nforcement), a unified single-stage framework that
mirrors radiologists' coarse-to-fine reasoning through a
clinically-driven three-level taxonomy of 9 anatomical regions,
17 sub-regions, and 28 pathological findings.
\modelname{} jointly optimizes multi-level supervision,
cross-level probability alignment, and a hierarchy-violation
penalty within a shared Vision Transformer backbone.
This ensures that fine-grained findings are anatomically
supported by their coarser-level context rather than predicted
in isolation.
Experiments demonstrate that \modelname{} outperforms flat and
hierarchical baselines across all levels while achieving superior
probabilistic hierarchy consistency, with level-wise attention
maps confirming anatomically grounded predictions.
Code is available at: \url{https://github.com/yejix-ai/CHASE}.


\end{abstract}

\section{Introduction}
Chest X-ray (CXR) interpretation is an inherently hierarchical,
coarse-to-fine process~\citep{chen2019deep, chen2020deep,
hanif2025enhancing, levatic2015importance, noor2024consistency}.
Radiologists first localize abnormalities within broad anatomical
systems before refining observations into specific sub-regions and
findings~\citep{nobel2022structured, yan2024ahive, moon2025lunguage}—a
flow that is clinically critical, as visually similar densities
(e.g., pneumonia vs.\ pleural effusion) require fundamentally
different management based on anatomical
context~\citep{klein2019systematic, kitazono2010differentiation,
rabaey2025modeling}.

Recent anatomy-aware approaches~\citep{agu2021anaxnet, yu2022anatomy,
muller2023anatomy} capture spatial dependencies but remain
detection-centric two-stage pipelines that classify findings
independently, without enforcing structural consistency across levels.
General hierarchical methods~\citep{park2024visually,
wehrmann2018hierarchical, giunchiglia2020coherent} enforce label
coherence via regularization, but are ill-suited to the multi-label,
multi-granularity nature of medical imaging where a single radiograph
may exhibit findings across multiple anatomical systems simultaneously.
CXR-specific efforts~\citep{chen2020deep, pham2021interpreting} address this partially via conditional probability training, yet remain confined to relatively shallow, disease-centric taxonomies with a limited set of 14 abnormalities.
Consequently, current systems frequently produce coarse-to-fine
contradictions that undermine clinical
reliability~\citep{asadi2025clinically, noor2024consistency}.

We argue these failures reflect the difficulty of learning reliable
coarse-to-fine dependencies without explicit cross-level
coherence~\citep{levatic2015importance, chen2020deep}---evidenced
in our ablation, where removing hierarchy-enforcing regularizers
reduces probabilistic consistency by over 9 points
(Section~\ref{sec:ablation}).
To address this, we propose \modelname{} (\textbf{C}lassification
with \textbf{H}ierarchical \textbf{A}nalysis and \textbf{S}tructured
\textbf{E}nforcement), a unified single-stage framework grounded in
a three-level taxonomy (9 regions $\rightarrow$ 17 sub-regions
$\rightarrow$ 28 findings) mirroring the coarse-to-fine clinical
flow. Within a shared Vision Transformer backbone, \modelname{}
jointly optimizes multi-level supervision, cross-level probability
alignment, and a hierarchy-violation penalty, structurally enforcing
that fine-grained predictions remain anatomically grounded in their
coarser-level context. Our contributions are as follows:
% \begin{itemize}
%     \item \textbf{Clinically-grounded taxonomy:} A three-level
%     hierarchy (9 regions, 17 sub-regions, 28 findings) mirroring
%     radiologists' coarse-to-fine reading order.
%     \item \textbf{Unified training objective:} A joint loss combining
%     multi-level supervision, Global KL Alignment, and a
%     Hierarchy-Violation Penalty for cross-level probabilistic
%     consistency.
%     \item \textbf{Empirical validation:} \modelname{} outperforms
%     baselines across all levels in AUC, achieves the highest
%     probabilistic consistency (74.26\%), and is validated through
%     localization analysis confirming anatomical grounding.
% \end{itemize}
\begin{itemize}
    \item \textbf{Clinically-grounded taxonomy:} A three-level hierarchy 
    (9 regions $\rightarrow$ 17 sub-regions $\rightarrow$ 28 findings) 
    mirroring radiologists' coarse-to-fine reading order, which induces 
    197 anatomically valid Top–Mid–Leaf paths.
    
    \item \textbf{Unified training objective:} A joint loss combining
    multi-level supervision, Global KL Alignment, and a
    Hierarchy-Violation Penalty for cross-level probabilistic
    consistency.
    
    \item \textbf{Empirical validation:} \modelname{} outperforms
    baselines across all levels in AUC, achieves the highest
    probabilistic consistency (74.26\%), and is validated through
    localization analysis confirming anatomical grounding.
\end{itemize}
\section{Method}
\label{sec:method}

\begin{figure*}[ht]
    \centering
    \includegraphics[width=\textwidth]{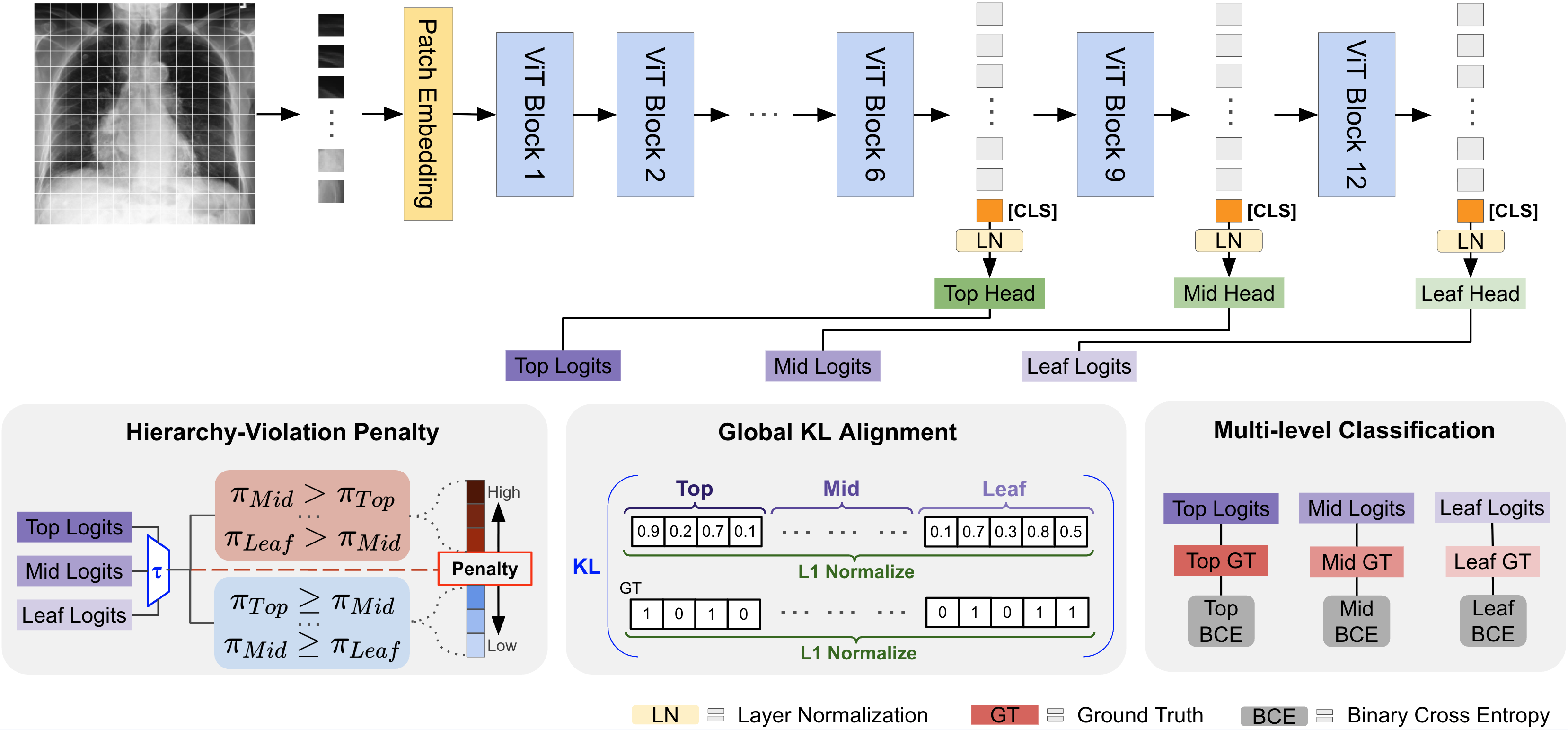}
    \caption{\footnotesize \textbf{Overview of the proposed hierarchical classification framework.}
    The model (top) adopts a ViT backbone with level-specific heads for \textit{Top}, \textit{Mid}, and \textit{Leaf} attached to progressively deeper transformer blocks. 
    Training uses a unified objective (bottom) comprising: (left) a \textbf{Hierarchy-Violation} penalty that suppresses parent--child contradictions; (center) a \textbf{Global KL Alignment} that couples all hierarchy levels via L1-normalized probability distributions; and (right) a \textbf{Multi-level Classification} loss that applies Binary Cross Entropy supervision at each granularity.}
    \label{fig:framework}
\end{figure*}

We present \modelname{}, a single-stage framework for fine-grained, clinically grounded multi-level classification from chest radiographs, allowing multiple labels to be predicted at each hierarchy level. As illustrated in Figure~\ref{fig:framework}, \modelname{} is built on (i) a clinically grounded three-level hierarchy and (ii) a unified training objective that enforces cross-level coherence via three complementary loss components.

\subsection{Clinically Grounded Three-Level Hierarchy}
\label{sec:method_hierarchy}
We structure the label space into a three-level hierarchy that aligns with radiologists' coarse-to-fine localization process (Table~\ref{tab:label_hierarchy_vocab}). Level 1 (\textbf{Top: Region}) comprises $C_1 = 9$ broad thoracic systems, establishing a global anatomical framework. Level 2 (\textbf{Mid: Sub-region \& Laterality}) refines these into $C_2 = 17$ precise zones, differentiating spatially adjacent structures and incorporating laterality. Level 3 (\textbf{Leaf: Finding \& Disease}) encompasses $C_3 = 28$ pathological labels, each linked to anatomically feasible parent locations.

\paragraph{Valid hierarchical relations.} 
Let $\mathcal{E}_{12} \subseteq \{1, \dots, C_1\} \times \{1, \dots, C_2\}$ represent the valid Region--Sub-region relations, $\mathcal{E}_{23} \subseteq \{1, \dots, C_2\} \times \{1, \dots, C_3\}$ denote the admissible Sub-region--Finding relations, and $\mathcal{E}_{13} \subseteq \{1, \dots, C_1\} \times \{1, \dots, C_3\}$ denote the induced Region--Finding relations along valid coarse-to-fine paths. Together, they define the set of anatomically valid hierarchical relations that serve as structural priors for cross-level coherence during training:
\begin{equation}
    \mathcal{E} = \mathcal{E}_{12} \cup \mathcal{E}_{23} \cup \mathcal{E}_{13}.
\end{equation}
The 9 Region, 17 Sub-region, and 28 Finding labels induce 197 anatomically valid \textit{Top}--\textit{Mid}--\textit{Leaf} paths.
\begin{table*}[t]
\centering
\caption{\footnotesize Three-level label hierarchy used in our study. Top captures coarse anatomical systems, Mid refines to sub-region and laterality, and Leaf defines specific findings.}
\label{tab:label_hierarchy_vocab}
\scriptsize
\setlength{\tabcolsep}{4pt}
\renewcommand{\arraystretch}{1.15}
\begin{tabular}{p{0.10\textwidth} p{0.88\textwidth}}
\toprule
\textbf{Level} & \textbf{Label vocabulary} \\
\midrule
\textbf{Top} &
\texttt{No Finding}, \texttt{lung}, \texttt{heart}, \texttt{mediastinum}, \texttt{trachea}, \texttt{spine}, \texttt{abdomen}, \texttt{clavicle}, \texttt{aorta} \\
\midrule
\textbf{Mid} &
\texttt{abdomen\_bilateral\_general}, \texttt{aorta\_bilateral\_general}, \texttt{clavicle\_left\_general}, \texttt{clavicle\_right\_general}, \texttt{heart\_bilateral\_general}, \texttt{lung\_left\_general}, \texttt{lung\_left\_lower}, \texttt{lung\_left\_mid}, \texttt{lung\_left\_upper}, \texttt{lung\_right\_general}, \texttt{lung\_right\_lower}, \texttt{lung\_right\_mid}, \texttt{lung\_right\_upper}, \texttt{mediastinum\_bilateral\_general}, \texttt{mediastinum\_bilateral\_upper}, \texttt{spine\_bilateral\_general}, \texttt{trachea\_bilateral\_general} \\
\midrule
\textbf{Leaf} &
\texttt{alveolar hemorrhage}, \texttt{aspiration}, \texttt{copd/emphysema}, \texttt{fluid overload/heart failure}, \texttt{goiter}, \texttt{granulomatous disease}, \texttt{interstitial lung disease}, \texttt{lung cancer}, \texttt{pericardial effusion}, \texttt{pneumonia}, \texttt{aorta changes}, \texttt{atelectasis}, \texttt{bone changes}, \texttt{bronchiectasis}, \texttt{cardiac changes}, \texttt{consolidation}, \texttt{cyst/bullae}, \texttt{hernia}, \texttt{hyperaeration}, \texttt{ild pattern}, \texttt{lung opacity}, \texttt{mass/nodule}, \texttt{mediastinal changes}, \texttt{pleural effusion}, \texttt{pleural scarring}, \texttt{pneumothorax}, \texttt{pulmonary edema}, \texttt{vascular changes} \\
\bottomrule
\end{tabular}
\end{table*}

\subsection{\modelname{} Architecture}
\label{sec:method_model}

\modelname{} adopts a Vision Transformer as its shared
backbone (Figure~\ref{fig:framework}).
Given an input radiograph $x$, level-specific linear heads are attached to
\texttt{[CLS]} embeddings extracted from blocks $k \in \{6, 9, 12\}$,
corresponding to the Top, Mid, and Leaf levels respectively:
\begin{equation}
\mathbf{z}_{\ell} = f_{\ell}\!\left(
    \text{LN}(\mathbf{h}^{(k_\ell)}_{\texttt{cls}})
\right), \quad \ell \in \{1, 2, 3\},
\label{eq:layerwise_heads}
\end{equation}
where $f_{\ell}$ is a linear projection and LN denotes Layer
Normalization. 
Unlike CNNs, ViTs exhibit uniform representations across layers with
global information accessible from early blocks~\citep{raghu2021vision},
meaning a coarse-to-fine hierarchy does \emph{not} emerge naturally
with depth—making explicit multi-depth supervision a necessity.
We attach heads at blocks 6, 9, and 12 (lower, middle, and final thirds of the 12-block backbone) to balance discriminative semantics and inter-level separation. Shallower representations, such as blocks 1/2/12, may better preserve level-specific separation but are less semantically discriminative, whereas later representations, such as blocks 5/11/12, provide stronger semantics but can make hierarchy-level predictions less distinct. This progressive placement provides granularity-specific supervision for regions, subregions, and findings within a shared encoder, while the core novelty of \modelname{} lies in the joint training objective described below.
%짧은버전
% We attach heads at blocks 6, 9, and 12 (lower, middle, and final thirds of the 12-block backbone) to balance discriminative semantics and inter-level separation. This progressive placement allows regions, subregions, and findings to be supervised at increasingly semantic stages while avoiding both overly shallow features and overly collapsed late representations. The core novelty of CHASE lies in the joint training objective, described below.

\subsubsection{Training Objective}
\label{sec:method_loss}

The model is trained to achieve per-level accuracy while maintaining
cross-level coherence by minimizing:
\begin{equation}
    \mathcal{L} = \mathcal{L}_{\text{cls}}
                + \alpha \, \mathcal{L}_{\text{KL}}
                + \beta \, \mathcal{L}_{\text{HVP}},
    \label{eq:total_loss}
\end{equation}
where $\mathcal{L}_{\text{cls}}$ provides per-level supervision,
$\mathcal{L}_{\text{KL}}$ couples all levels by global distribution alignment,
and $\mathcal{L}_{\text{HVP}}$ enforces hierarchy consistency under the valid hierarchical relations $\mathcal{E}$.
The weights $(\alpha, \beta) = (0.1, 0.3)$ are selected via ablation (Section~\ref{sec:ablation}).

\paragraph{(i) Multi-level Classification}
\label{sec:method_cls}

Let $\mathbf{t}_{\ell} \in \{0,1\}^{C_{\ell}}$ represent the multi-hot
ground-truth labels for level $\ell$.
We impose independent binary cross-entropy (BCE) over each label dimension at
every hierarchy level:
\begin{equation}
    \mathcal{L}_{\text{cls}} = \sum_{\ell=1}^{3}
    \sum_{j=1}^{C_{\ell}} \left[
    - t_{\ell,j} \log \hat{p}_{\ell,j}
    - (1 - t_{\ell,j}) \log (1 - \hat{p}_{\ell,j})
    \right],
\end{equation}
where $\hat{p}_{\ell,j}$ is the predicted probability
for the $j$-th label at level $\ell$.
This ensures that all three levels develop discriminative representations directly from their own supervision signals.

\paragraph{(ii) Global KL Alignment}
\label{sec:method_kl}

While $\mathcal{L}_{\text{cls}}$ optimizes each level independently,
cross-level coherence requires that the joint predicted distribution remains
compatible with the ground-truth active label set.
We enforce this by aligning the concatenated prediction distribution with
the ground-truth via KL divergence. Concatenating logits and targets yields
$\mathbf{z} = [\mathbf{z}_{1}; \mathbf{z}_{2}; \mathbf{z}_{3}]$ and
$\mathbf{t} = [\mathbf{t}_{1}; \mathbf{t}_{2}; \mathbf{t}_{3}]$.
Probabilities are derived via temperature-scaled sigmoid followed by
L1 normalization:
\begin{equation}
    \mathbf{p} = \frac{\sigma(\mathbf{z}/\tau)}
                      {\|\sigma(\mathbf{z}/\tau)\|_{1} + \varepsilon},
    \qquad
    \mathbf{q} = \frac{\mathbf{t}}{\|\mathbf{t}\|_{1} + \varepsilon},
\end{equation}
with temperature $\tau = 1$ fixed across all experiments.
The KL term is applied to samples with at least one positive label:
\begin{equation}
    \mathcal{L}_{\text{KL}} = D_{\text{KL}}(\mathbf{q} \, \| \, \mathbf{p}).
\end{equation}
This regularizer discourages probability mass from concentrating on
fine-grained findings that lack support from coarser anatomical
contexts.\footnote{For normal studies (\textit{No Finding} only),
$\mathbf{q}$ degenerates to a coarsest-level one-hot vector,
naturally suppressing all fine-grained activations—the clinically
desired behaviour for normal radiographs.}

\paragraph{(iii) Hierarchy-Violation Penalty}
\label{sec:method_penalty}

To suppress parent-child contradictions, we introduce a differentiable
regularizer that enforces probabilistic monotonicity across hierarchy levels
(Figure~\ref{fig:framework}).
Let $\boldsymbol{\pi}_{\ell} = \sigma(\mathbf{z}_{\ell})$ denote the
predicted probabilities at level $\ell$.
Clinically, a finding should not be predicted with higher confidence than
its anatomical parent—i.e., valid predictions satisfy
$\pi_{\text{Top}} \geq \pi_{\text{Mid}} \geq \pi_{\text{Leaf}}$ along any
coarse-to-fine path.
For each admissible relation $(p, c) \in \mathcal{E}$, we penalize
violations of this monotonicity:
\begin{equation}
    \mathcal{L}_{\text{HVP}} =
    \sum_{(p,c) \in \mathcal{E}} \pi_{c} \,(1 - \pi_{p}).
\label{eq:penalty}
\end{equation}
The penalty is \emph{asymmetric} by design: it targets the clinically
critical error of activating a fine-grained finding without broader
anatomical support ($\pi_c > \pi_p$), while leaving the reverse
unpunished—reflecting that a radiologist may flag a broad region as
abnormal before committing to a specific finding.
\section{Results}

\subsection{Experimental Setup}
\label{sec:exp_setup}

\paragraph{Dataset and implementation details.}
We derive a three-level hierarchy (Top/Mid\\
/Leaf) from Chest ImaGenome \citep{chest_imagenome} using MIMIC-CXR \citep{mimic-cxr}. Following the official splits and keeping all frontal views, we obtain 213,361/1,733/3,041 (train/val/test) studies, corresponding to 237,070/1,959/3,403 frontal images after retaining all available frontal images per study. All models are evaluated on a unified 54-label space, with metrics reported at the image level ($N=3,403$). To ensure performance gains reflect our proposed designs rather than latent representation quality, all models share an identical EVA-X~\citep{yao2025evax} backbone, providing domain-specific visual features via a ViT-based foundation model. We fine-tune end-to-end for 20 epochs with $(\alpha,\beta)=(0.1,0.3)$\footnote{Hyperparameters: batch size 512, AdamW (LR=$1\times10^{-5}$, WD=0.05), 2-epoch warmup. Results are reported as mean$\pm$std over 1,000 bootstrap seeds.}.

\begin{table}[t]
\centering
\caption{\footnotesize
Main performance comparison on hierarchical classification.
\textbf{Single-level} models are trained and evaluated on specific
hierarchy levels, while \textbf{Multi-level} models handle all levels
simultaneously. Best results are highlighted in \textbf{bold}.
}
\label{table:main_results}
\resizebox{\columnwidth}{!}{%
  \begin{tabular}{l c ccc ccc cc}
  \toprule
  \multirow{2}{*}{\textbf{Model}} &
  \multirow{2}{*}{\textbf{\makecell{\# of\\Classes}}} &
  \multicolumn{3}{c}{\textbf{AUC} ($\uparrow$)} &
  \multicolumn{3}{c}{\textbf{ACC} ($\uparrow$)} &
  \multicolumn{2}{c}{\textbf{Consis.} ($\uparrow$)} \\
  \cmidrule(lr){3-5} \cmidrule(lr){6-8} \cmidrule(lr){9-10}
  & & \textbf{Top} & \textbf{Mid} & \textbf{Leaf}
    & \textbf{Top} & \textbf{Mid} & \textbf{Leaf}
    & \textbf{Det.} & \textbf{Prob.} \\
  \midrule
  \multicolumn{10}{l}{\textit{Single-level Evaluation (Top / Mid / Leaf)}} \\
  \addlinespace[2pt]
  Flat-ViT (Top)  & 9  &
    0.847{\scriptsize$\pm$.009} & -- & -- &
    91.81{\scriptsize$\pm$.184} & -- & -- & -- & -- \\
  Flat-ViT (Mid)  & 17 &
    -- & 0.805{\scriptsize$\pm$.007} & -- &
    -- & 87.63{\scriptsize$\pm$.164} & -- & -- & -- \\
  Flat-ViT (Leaf) & 28 &
    -- & -- & 0.847{\scriptsize$\pm$.004} &
    -- & -- & 91.96{\scriptsize$\pm$.094} & -- & -- \\
  \addlinespace[2pt]
  \midrule
  \multicolumn{10}{l}{\textit{Multi-level Evaluation (Top \& Mid \& Leaf)}} \\
  \addlinespace[2pt]
  Flat-ViT  & 54 &
    0.732{\scriptsize$\pm$.012} & 0.702{\scriptsize$\pm$.009} & 0.770{\scriptsize$\pm$.005} &
    91.51{\scriptsize$\pm$.180} & 86.61{\scriptsize$\pm$.174} & 91.59{\scriptsize$\pm$.093} &
    92.77{\scriptsize$\pm$.047} & 70.16{\scriptsize$\pm$.065} \\
  Hier-ViT  & 54 &
    0.799{\scriptsize$\pm$.010} & 0.764{\scriptsize$\pm$.007} & 0.774{\scriptsize$\pm$.005} &
    91.60{\scriptsize$\pm$.183} & 87.25{\scriptsize$\pm$.166} & 91.64{\scriptsize$\pm$.090} &
    93.46{\scriptsize$\pm$.055} & 71.12{\scriptsize$\pm$.101} \\
  H-CAST    & 54 &
    0.785{\scriptsize$\pm$.011} & 0.755{\scriptsize$\pm$.007} & 0.792{\scriptsize$\pm$.005} &
    90.81{\scriptsize$\pm$.191} & 87.27{\scriptsize$\pm$.165} & 91.64{\scriptsize$\pm$.093} &
    93.42{\scriptsize$\pm$.055} & 72.10{\scriptsize$\pm$.115} \\
  \addlinespace[2pt]
  \midrule
  \rowcolor{gray!12}
  \textbf{Ours (Full)} & 54 &
    \textbf{0.826}{\scriptsize$\pm$.009} &
    \textbf{0.803}{\scriptsize$\pm$.007} &
    \textbf{0.848}{\scriptsize$\pm$.004} &
    \textbf{91.63}{\scriptsize$\pm$.183} &
    \textbf{87.77}{\scriptsize$\pm$.162} &
    \textbf{91.83}{\scriptsize$\pm$.096} &
    \textbf{94.96}{\scriptsize$\pm$.057} &
    \textbf{74.26}{\scriptsize$\pm$.115} \\
  \bottomrule
  \end{tabular}
}
\end{table}

\paragraph{Metrics.}
We report per-level macro-averaged AUC and accuracy (ACC). To quantify cross-level coherence beyond independent level-wise metrics, we introduce hierarchy consistency (Consis.) metric. As defined in Sec.~\ref{sec:method_hierarchy}, the valid relations $\mathcal{E}$ specify anatomically admissible Top$\rightarrow$Mid$\rightarrow$Leaf paths. Consis. is the fraction of samples whose predictions are monotone along every such path. 
Let $\pi^{(i)}_{\ell,j}$ be the predicted probability of label $j$ at level $\ell$ for sample $i$, and set $g^{(i)}_{\ell,j}=\mathbf{1}[\pi^{(i)}_{\ell,j}\ge\tau]$ (Consis.-Det.) or $g^{(i)}_{\ell,j}=\pi^{(i)}_{\ell,j}$ (Consis.-Prob.). We define $\mathrm{Consis.}(\tau)=\frac{1}{N}\sum_{i=1}^{N}\mathbf{1}\!\left[g^{(i)}_{\mathrm{Top},j}\ge g^{(i)}_{\mathrm{Mid},k}\ge g^{(i)}_{\mathrm{Leaf},l},\ \forall (j,k)\in\mathcal{E}_{12},\ (k,l)\in\mathcal{E}_{23}\right]$ and report it at $\tau=0.5$. Consis.-Prob.\ is a strictly more demanding criterion, as it requires probabilistic monotonicity to hold continuously rather than only after hard thresholding.

\paragraph{Baselines.}
% Direct comparison with prior hierarchical CXR studies is inherently limited, as those models rely on proprietary label systems, multi-stage pipelines, or datasets incompatible with our unified 54-label evaluation protocol—making faithful reimplementation 
% intractable. We therefore focus on modern, reproducible baselines 
% that can be rigorously evaluated under identical conditions. 
% Specifically, we compare \modelname{} against \textit{Flat-ViT}, which 
% predicts all levels simultaneously without any hierarchical 
% inductive bias, and train independent single-level versions 
% (\textit{Flat-ViT (Top/Mid/Leaf)}) to serve as per-level 
% performance upper bounds. We further include \textit{Hier-ViT}~\citep{park2024visually} 
% and \textit{H-CAST}~\citep{park2024visually}, two state-of-the-art hierarchical classifiers from the recent general-domain literature, adapted to our label space. Importantly, both follow a fine-to-coarse aggregation strategy—the reverse of our clinically motivated top-down approach—providing a direct ablative contrast that isolates the diagnostic benefit of aligning prediction order with radiologists' coarse-to-fine reasoning.
Prior hierarchical CXR methods~\citep{chen2020deep, pham2021interpreting, agu2021anaxnet, muller2023anatomy} rely on mutually incompatible label spaces, making cross-study AUC comparison statistically ill-defined.
We instead establish a reproducible 54-label protocol derived from Chest ImaGenome~\citep{chest_imagenome}, with all baselines evaluated under identical conditions. We compare \modelname{} against \textit{Flat-ViT} (all 54 labels, no hierarchical bias) and single-level \textit{Flat-ViT (Top/Mid/Leaf)} as oracle performance ceilings—the key measure being how much \modelname{} narrows the gap relative to multi-level baselines. We further include \textit{Hier-ViT} and \textit{H-CAST}~\citep{park2024visually}, adapted to our label space by replacing classification heads with level-specific projections while preserving all hierarchical inductive biases.
Critically, both follow a \emph{fine-to-coarse} aggregation strategy—the reverse of our top-down approach—providing a direct ablative contrast for the benefit of coarse-to-fine alignment.

\subsection{Hierarchical Classification Performance}
\label{sec:main_results}
Table~\ref{table:main_results} compares \modelname{} against both
single-level and multi-level models.
\modelname{} achieves the strongest performance across all levels
(Top/Mid/Leaf AUC: 0.826\\/0.803/0.848), matching or exceeding the
oracle single-level ceilings demonstrating that hierarchical training
benefits rather than compromises fine-grained classification.
The multi-level Flat-ViT shows a large AUC drop (0.847$\to$0.732 at
Top), confirming that naïve joint training is insufficient; Hier-ViT
and H-CAST recover some performance but remain below \modelname{},
particularly at Mid and Leaf levels where fine-to-coarse error
propagation is most pronounced.
% Beyond accuracy, \modelname{} achieves the highest probabilistic
% consistency (Consis.-Prob.: 74.26\%), outperforming all multi-level
% baselines.
Beyond level-wise AUC, \modelname{} achieves the highest probabilistic consistency (Consis.-Prob.: 74.26\%) over the 197 anatomically valid \textit{Top}--\textit{Mid}--\textit{Leaf} paths, confirming improved cross-level coherence across the full hierarchy.

% Note that the HVP-only ablation (Table~\ref{table:ablation}) attains higher Consis.-Prob.\ (78.72\%) but at the cost of Leaf AUC (0.837); \modelname{} achieves the most favourable accuracy--consistency trade-off overall.

\subsection{Ablation Studies}
\label{sec:ablation}

\begin{table}[t]
\centering
\caption{\footnotesize
Weight ablation of Global KL ($\alpha$) and Hierarchy-Violation
Penalty ($\beta$). \textit{Consis.-Det.} and \textit{Consis.-Prob.}
denote consistency at $\tau{=}0.5$ and probabilistic monotonicity,
respectively. Best and second-best results are in \textbf{bold} and
\underline{underlined}.
}
\label{tab:weight_ablation}
\setlength{\tabcolsep}{4.6pt}
\renewcommand{\arraystretch}{1.08}
\resizebox{\columnwidth}{!}{
\begin{tabular}{l cc ccc ccc cc}
  \toprule
  \multirow{2}{*}{\textbf{Configuration}} &
  \multicolumn{2}{c}{\textbf{Weights}} &
  \multicolumn{3}{c}{\textbf{AUC} ($\uparrow$)} &
  \multicolumn{3}{c}{\textbf{ACC} ($\uparrow$)} &
  \multicolumn{2}{c}{\textbf{Consis.} ($\uparrow$)} \\
  \cmidrule(lr){2-3} \cmidrule(lr){4-6}
  \cmidrule(lr){7-9} \cmidrule(lr){10-11}
  & $\alpha$ & $\beta$ &
  \textbf{Top} & \textbf{Mid} & \textbf{Leaf} &
  \textbf{Top} & \textbf{Mid} & \textbf{Leaf} &
  \textbf{Det.} & \textbf{Prob.} \\
  \midrule
  \multicolumn{11}{l}{\textit{Anchor configurations}} \\
  \addlinespace[2pt]
  No reg.
    & 0.0 & 0.0
    & 0.763{\scriptsize$\pm$.012} & 0.752{\scriptsize$\pm$.008} & 0.791{\scriptsize$\pm$.007}
    & 91.43{\scriptsize$\pm$.184} & 87.33{\scriptsize$\pm$.165} & 91.95{\scriptsize$\pm$.094}
    & 91.96{\scriptsize$\pm$.078} & 64.89{\scriptsize$\pm$.119} \\
  KL only
    & 1.0 & 0.0
    & 0.819{\scriptsize$\pm$.009} & 0.803{\scriptsize$\pm$.007} & \underline{0.848}{\scriptsize$\pm$.004}
    & 91.01{\scriptsize$\pm$.191} & 87.70{\scriptsize$\pm$.162} & \textbf{92.10}{\scriptsize$\pm$.092}
    & 91.88{\scriptsize$\pm$.088} & 65.26{\scriptsize$\pm$.162} \\
  HVP only
    & 0.0 & 1.0
    & \underline{0.825}{\scriptsize$\pm$.009} & 0.801{\scriptsize$\pm$.007} & 0.837{\scriptsize$\pm$.004}
    & 91.53{\scriptsize$\pm$.188} & 87.50{\scriptsize$\pm$.162} & 90.40{\scriptsize$\pm$.106}
    & \textbf{97.57}{\scriptsize$\pm$.048} & \textbf{78.72}{\scriptsize$\pm$.108} \\
  Both (1.0, 1.0)
    & 1.0 & 1.0
    & 0.801{\scriptsize$\pm$.011} & 0.800{\scriptsize$\pm$.007} & 0.843{\scriptsize$\pm$.004}
    & 90.99{\scriptsize$\pm$.184} & 87.71{\scriptsize$\pm$.165} & 91.69{\scriptsize$\pm$.098}
    & 95.64{\scriptsize$\pm$.051} & \underline{76.63}{\scriptsize$\pm$.096} \\
  Both (0.5, 0.5)
    & 0.5 & 0.5
    & 0.809{\scriptsize$\pm$.010} & 0.801{\scriptsize$\pm$.007} & 0.846{\scriptsize$\pm$.004}
    & 91.32{\scriptsize$\pm$.181} & 87.78{\scriptsize$\pm$.164} & 91.82{\scriptsize$\pm$.096}
    & 95.13{\scriptsize$\pm$.054} & 75.67{\scriptsize$\pm$.098} \\
  \addlinespace[2pt]
  \midrule
  \multicolumn{11}{l}{\textit{$\alpha\downarrow$\,$\beta\uparrow$ progression}} \\
  \addlinespace[2pt]
  $\alpha\downarrow$ $\beta\uparrow$-1
    & 0.5 & 0.1
    & 0.822{\scriptsize$\pm$.009} & \underline{0.804}{\scriptsize$\pm$.007} & \underline{0.848}{\scriptsize$\pm$.004}
    & 91.37{\scriptsize$\pm$.184} & \underline{87.79}{\scriptsize$\pm$.164} & \underline{92.09}{\scriptsize$\pm$.092}
    & 93.00{\scriptsize$\pm$.066} & 70.45{\scriptsize$\pm$.136} \\
  $\alpha\downarrow$ $\beta\uparrow$-2
    & 0.3 & 0.1
    & \underline{0.825}{\scriptsize$\pm$.008} & \textbf{0.805}{\scriptsize$\pm$.007} & \textbf{0.849}{\scriptsize$\pm$.004}
    & 91.45{\scriptsize$\pm$.185} & \textbf{87.81}{\scriptsize$\pm$.161} & \underline{92.09}{\scriptsize$\pm$.093}
    & 93.12{\scriptsize$\pm$.065} & 70.49{\scriptsize$\pm$.135} \\
  \addlinespace[2pt]
  \rowcolor{gray!12}
  \modelname{} ($\alpha\downarrow$ $\beta\uparrow$-3)
    & \textbf{0.1} & \textbf{0.3}
    & \textbf{0.826}{\scriptsize$\pm$.009} & 0.803{\scriptsize$\pm$.007} & \underline{0.848}{\scriptsize$\pm$.004}
    & \textbf{91.63}{\scriptsize$\pm$.183} & 87.77{\scriptsize$\pm$.162} & 91.83{\scriptsize$\pm$.096}
    & 94.96{\scriptsize$\pm$.057} & 74.26{\scriptsize$\pm$.115} \\
  \addlinespace[2pt]
  $\alpha\downarrow$ $\beta\uparrow$-4
    & 0.1 & 0.5
    & 0.822{\scriptsize$\pm$.009} & 0.802{\scriptsize$\pm$.007} & 0.846{\scriptsize$\pm$.004}
    & \textbf{91.63}{\scriptsize$\pm$.184} & 87.65{\scriptsize$\pm$.163} & 91.42{\scriptsize$\pm$.098}
    & \underline{95.95}{\scriptsize$\pm$.055} & 75.30{\scriptsize$\pm$.112} \\
  \bottomrule
\end{tabular}%
}
\end{table}
% Table~\ref{tab:weight_ablation} shows how the Global KL ($\alpha$) and Hierarchy-Violation Penalty ($\beta$) terms contribute to the model's performance. Without any regularization ($\alpha{=}\beta{=}0$), the model fails to maintain structural logic, resulting in low consistency (64.89). We observe a clear trade-off between discriminative power and hierarchical coherence: adding KL alignment alone ($\alpha{=}1$) improves AUC and accuracy across all levels (e.g., 0.848 Leaf AUC) but provides little gain in coherence (65.26), confirming that per-level performance does not guarantee a logical hierarchy. Conversely, using the HVP alone ($\beta{=}1$) maximizes consistency (78.72) but degrades Leaf-level discrimination (0.837 AUC), as strict monotonicity can over-constrain fine-grained predictions. We find that our final setting ($\alpha{=}0.1, \beta{=}0.3$) achieves the most favorable balance, yielding the highest Top AUC (0.826) and strong overall consistency (74.26). Further increasing the penalty (e.g., $\beta{=}0.5$) leads to diminishing returns, where accuracy drops without significant gains in coherence.
Table~\ref{tab:weight_ablation} examines the individual and joint effects of the Global KL weight ($\alpha$) and Hierarchy-Violation Penalty ($\beta$). Without regularization ($\alpha{=}\beta{=}0$), the model fails to maintain structural logic, yielding low consistency (64.89). KL alignment serves as a calibration signal: aligning predicted distributions with labels improves per-level AUC (e.g., Leaf AUC: 0.791~$\to$~0.848) but yields negligible consistency gains (65.26). In contrast, HVP directly enforces monotonicity; HVP-only ($\beta{=}1.0$) maximizes consistency (78.72) but degrades Leaf AUC (0.837) by over-constraining fine-grained predictions. Our final setting ($\alpha{=}0.1, \beta{=}0.3$) achieves the most favorable balance, yielding the highest Top AUC (0.826) and strong consistency (74.26) without accuracy degradation. Further increasing $\beta$ (e.g., 0.5) leads to diminishing returns, where accuracy drops without commensurate gains in coherence.

\subsection{Hierarchy-Consistent Localization Results}
\label{sec:localization}

\begin{figure*}[t]
    \centering
    \includegraphics[width=\textwidth]{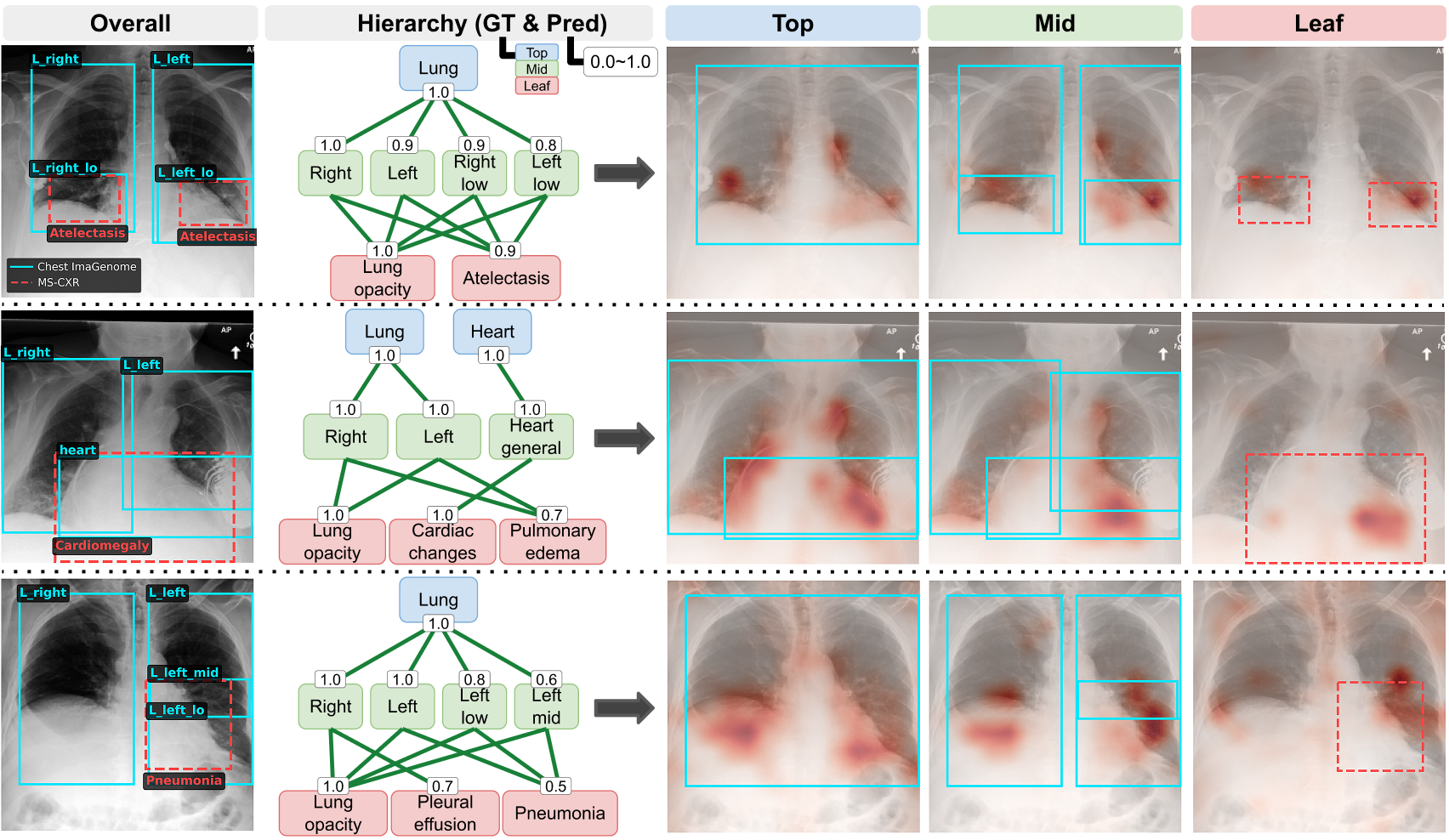}
    \caption{\footnotesize \textbf{Hierarchy-consistent localization via attention maps.}
    For each example, we show (left) the input image with reference boxes from Chest ImaGenome (cyan) and MS-CXR (red dashed), (middle) the hierarchy graph with ground-truth nodes and predicted probabilities, and (right) level-wise attention maps derived from ViT self-attention (\texttt{[CLS]}$\rightarrow$patch) at Blocks 6/9/12 for Top/Mid/Leaf, respectively.}
    \label{fig:figure3}
\end{figure*}

% Figure~\ref{fig:figure3} shows how \modelname{} grounds its predictions in anatomically meaningful evidence via attention mapss from Blocks 6, 9, and 12. Leaf-level attention is highly localized, closely matching disease-specific regions and reference bounding boxes. Moving to Mid and Top levels, the attention expands around the initial Leaf evidence to cover the predicted sub-regions and broader anatomical systems. 

Figure~\ref{fig:figure3} shows how \modelname{} grounds its
predictions in anatomically meaningful evidence via attention
maps from Blocks 6, 9, and 12. Leaf-level attention is highly
localized, closely matching disease-specific regions and
reference bounding boxes. Moving to Mid and Top levels,
attention expands to cover predicted sub-regions and broader
anatomical systems. This progression suggests that coarse
labels are supported by localized abnormal areas rather than
diffuse ``whole-organ'' cues. The hierarchy panel confirms
this probabilistic consistency: 
Top-level attention highlights the correct anatomical system,
Mid-level distributes across corresponding sub-regions, and
Leaf-level concentrates on the final finding.
This path-level evidence suggests practical clinical utility by making predictions anatomically auditable: radiologists can inspect whether a fine-grained finding is supported by the expected sub-region and broader anatomical system.
Collectively, these results confirm that \modelname{}
produces hierarchy-consistent predictions that follow an
anatomically coherent coarse-to-fine reasoning process.

% To quantitatively validate this spatial grounding, we evaluate attention localisation using the \textbf{Pointing Game} metric~\citep{zhang2016pointing}: a prediction is counted as correct if the peak attention location falls within the annotated bounding box from Chest ImaGenome or MS-CXR. % [FIX-⑨ 주의] 아래 수치는 실제 실험 결과로 교체 필요.
% %               현재는 placeholder (XX.X\%) 로 표기.
% \modelname{} achieves Pointing Game accuracy of XX.X\%/XX.X\%/XX.X\% at the Top/Mid/Leaf levels respectively, consistently outperforming the multi-level Flat-ViT baseline (XX.X\%/XX.X\%/XX.X\%).
\section{Conclusions}
We introduced \modelname{}, a single-stage hierarchical multi-label classifier that aligns chest X-ray interpretation with clinical coarse-to-fine reasoning. 
By attaching level-specific heads at multiple ViT depths, \modelname{} imposes explicit hierarchical supervision that aligns each block toward its assigned anatomical granularity level.
 To ensure cross-level coherence, we optimized a unified objective combining multi-level supervision with Global KL Alignment and a Hierarchy-Violation Penalty. Empirically, \modelname{} achieves strong performance across all levels while significantly improving probabilistic consistency. Attention maps further confirm that predictions progress from broad anatomical coverage to localized disease evidence, supporting the model's interpretability. These results demonstrate that clinically aligned hierarchical supervision improves the reliability of fine-grained CXR classification by promoting path-level anatomical coherence. Future work will focus on extending the taxonomy and evaluating robustness across diverse clinical settings.

\bibliographystyle{splncs04}
\bibliography{ref}
\end{document}